\documentclass{article}
\usepackage{spconf,amsmath,graphicx}
\usepackage{xcolor}
\usepackage{amsfonts}

\usepackage{multirow}
\usepackage{algorithm}
\usepackage{algorithmicx}
\usepackage{algpseudocode}
\usepackage{amsmath}

\usepackage{verbatim}

\usepackage{xcolor}
\usepackage{amsfonts}

\usepackage{multirow}
\usepackage{algorithm}
\usepackage{algorithmicx}
\usepackage{algpseudocode}
\usepackage{amsmath}

\title{Anomaly Detection by One Class Latent Regularized Networks}
\name{Chengwei Chen$^{1}$  \quad Pan Chen$^{2}$ \quad Haichuan Song$^{1}$ \quad Yiqing Tao$^{1}$ \quad Yuan Xie$^{1}$$^{\dagger}$ \quad Lizhuang Ma$^{1}$ \thanks{${\dagger}$ Yuan Xie (xieyuan8589@foxmail.com) is the correspongding author.}}

\address{$^{1}$ East China Normal University \\
	$^{2}$ Shanghai Jiao Tong University }
\begin{document}
%
\maketitle
\begin{abstract}

Anomaly detection is a fundamental problem in computer vision area with many real-world applications. Given a wide range of images belonging to the normal class, emerging from some distribution, the objective of this task is to construct the model to detect out-of-distribution images belonging to abnormal instances. Semi-supervised Generative Adversarial Networks (GAN)-based methods have been gaining popularity in anomaly detection task recently. However, the training process of GAN is still unstable and challenging. To solve these issues, a novel adversarial dual autoencoder network is proposed, in which the underlying structure of training data is not only captured in latent feature space, but also can be further restricted in the space of latent representation in a discriminant manner, leading to a more accurate detector. In addition, the auxiliary autoencoder regarded as a discriminator could obtain an more stable training process. Experiments show that our model achieves the state-of-the-art results on MNIST and CIFAR10 datasets as well as GTSRB stop signs dataset.
\end{abstract}

\section{Introduction}

With the growing popularity of self-drive \cite{kawamura1987motorized} and the widely usage of surveillance cameras \cite{nanri2005unsupervised}, it has become urgent to distinguish abnormal instances from normal ones. The anomaly detection task could be defined as follows: given a wide range of normal instances, one must determine whether the input data exhibits any irregularity. However, in the real world, the data resources are highly imbalanced towards samples of the normal class, whilst lacking in the samples of the abnormal class. In addition, it is impossible to take all types of abnormal data into account. \emph{Hence, the main challenge of this task is to exploit the real characteristic of normal samples only and detect the points that deviate from the normality.}

In the past, sparse representation \cite{cong2011sparse} and dictionary learning approaches \cite{lu2013abnormal} \cite{han2013online} make great achievement in abnormal detection task. Sparse representation is adopted to learn the dictionary of normal behaviors. During the testing period, the patterns which have large reconstruction errors are considered as anomalous behaviors. Actually, collecting annotations for training purpose is costly and time-consuming, especially for videos. Nanri \cite{nanri2005unsupervised} suggests that unsupervised anomaly detection techniques could uncover anomalies in an unlabeled test data. Unsupervised abnormal detection methods such as the One-class Support Vector Machines \cite{li2003improving} and Kernel Density Estimation \cite{latecki2007outlier} are widely used and effectively identify the abnormal samples. Nevertheless, this kind of methods suffer from sub-optimal issue, when they deal with the complex and high dimension datasets.

In recent researches, a lot of abnormal detection methods \cite{ravanbakhsh2017abnormal} \cite{akcay2018ganomaly} \cite{sabokrou2018adversarially} \cite{vasilev2018q} based on the Generative Adversarial Networks (GANs) are proposed. The GAN style architecture is adversarially trained under the semi-supervised learning framework, in which the real characteristic of target class is captured in the latent space. In the process of testing, the abnormal samples are regarded as the out-of-distributions samples that naturally exhibit a higher pixel-wise reconstruction error than normal samples. However, the real objective of the task is supposed to capture more separable latent features between normal samples and abnormal samples instead of minimizing the pixel-wise reconstruction error. In addition, conventional GAN style architectures always obtain blurry reconstructions because of the existing of multiple modes in the actual normal distribution. The blurriness falsifies reconstruction errors, which is disastrous for anomaly detection task. Moreover, the imbalance of capability between generator and discriminator leads to an unstable training process in generative adversarial network, which limits the capacity of GAN style architecture.

Motivated by the above limitations, we propose a novel adversarial dual autoencoder network under the semi-supervised learning framework. Compared with conventional GANs, a latent regularizer and an auxiliary autoencoder are adopted to the framework. The latent regularizer could make the latent features of target class more concentrated, which further enlarge the gap between the normal samples and abnormal samples. Besides, in conventional GAN style architecture, the generator and discriminator compete with each other. It will produce the imbalance of capability between these two subnetworks, leading to an unstable training process. To deal with this issue, the auxiliary autoencoder regarded as an another discriminator could obtain a better balance in the training process. In addition, only the normal samples are reconstructed well through the dual autoencoder framework.

\section{Proposed Method}

\begin{figure*}[h]
	\centering
	\includegraphics[width=0.78\linewidth]{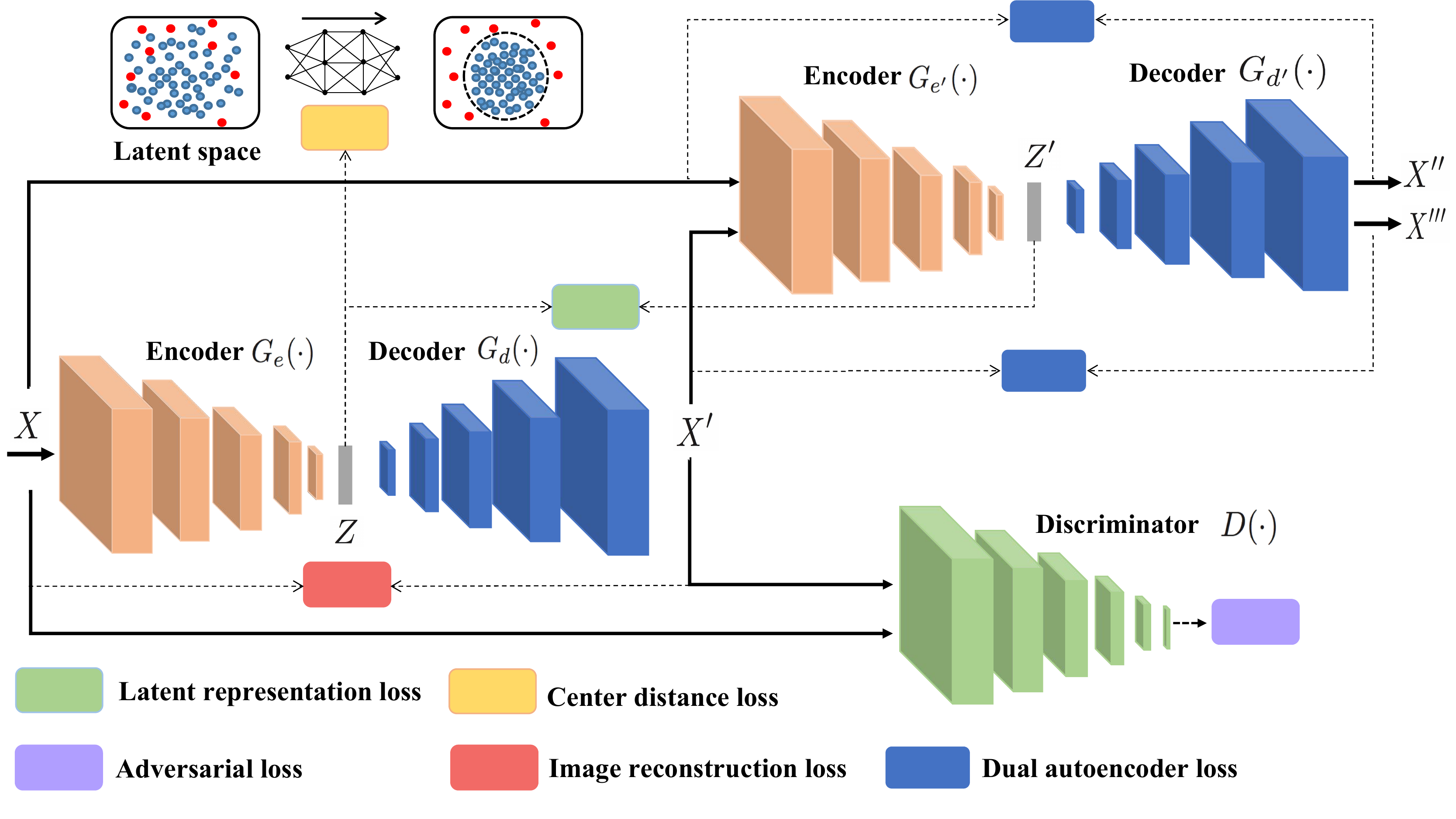}
	\caption{Our framework consists of two autoencoders, one discriminator and one latent regularizer.}
	\label{achieture}
\end{figure*}

In this part, we start by describing the details of the proposed network architecture, then depict each term in loss function.

\subsection{Network Architecture}

Proposed adversarial dual autoencoder network, which is shown in Fig. \ref{achieture}, consists of four components: two autoencoder networks, a discriminator and an one-class regularizer.

The main function of the first autoencoder regarded as generator is to reconstruct the input image to ``fool'' the discriminator. At the same time, the discriminator tries to distinguish between reconstruction image and original input image. Both of them compete with each other to obtain high-quality reconstruction images that even the discriminator could not be determined. The generator learns the real characteristic of target class by minimizing the pixel-wise error between original input image and reconstruction image.

To avoid being fooled by the generator, the discriminator tries to learn the real characteristic of the target class in the process of adversarial training and distinguishes original input image from generated image. During this period, the discriminator can help the generator to obtain more stable and robust parameters, which could also enlarge the gap between normal and abnormal samples.

The auxiliary autoencoder is structed as a discriminator, which has the same architecture as the generator but with different parametrization. The reason why we employ the auxiliary autoencoder can be concluded as follows: (1) Discriminator is usually a binary classifier as the component that we have described above, which distinguish original input images from generated images. Generator and discriminator are trained by competing with each other. However, it will produce the imbalance of capability between these two subnetworks, leading to an unstable training process. Some \cite{zhao2016energy} \cite{berthelot2017began} suggest that the auxiliary autoencoder could make a balance between the capability of generator and discriminator. (2) Only the normal images can be reconstructed well through adversarial dual autoencoder framework. To some extend, compared with conventional adversarial autoencoder network, proposed adversarial dual autoencoder framework would make the normalities and the anomalies more separable.

To further identify the latent space, the latent regularizer contracts the sphere by minimizing the mean distance of all latent features to the center. This center is calculated by the mean of the latent features that result from performing an initial forward pass on all training data. In order to map the latent feature to center $c$ as close as possible, the generator must capture the common factors of variation. Therefore, we design a latent regularizer by penalizing all samples based on the distance to the center in the latent space.

\subsection{Overall Loss Function}

In the training process, we define a total loss function in Eqn. \ref{loss-function} including five components, the adversarial loss, the image reconstruction loss, the center distance loss, the latent representation loss and the dual autoencoder loss, which can be formulated as :
\begin{equation}\label{loss-function}
\mathcal{L}=w_{i} \mathcal{L}_{irec} + w_{a} \mathcal{L}_{adv} +w_{z} \mathcal{L}_{zrec}+w_{c} \mathcal{L}_{c} + w_{d} \mathcal{L}_{dual},
\end{equation}
 where $w_{i}$, $w_{a}$, $w_{z}$, $w_{c}$, and $w_{d}$ are the weighting parameters balancing the impact of individual item to the overall object function. 

\textbf{Adversarial  loss:}
Adversarial loss is adopted to train the first generator $G$ and discriminator $D$. The discriminator tries to distinguish between generated images $G(\mathbf{x})$ and original input images $x$, while the first generator attempts to learn the real characteristic of the target class and generates the image to fool the discriminator. According to the previous work \cite{goodfellow2014generative}, this adversarial game could be formulated as:

\begin{equation}
	\label{eq:2}
\begin{aligned} \mathcal{L}_{a d v}=& \min _{G} \max _{D}\left(E_{\boldsymbol{x} \sim p_{\mathbf{x}}}[\log (D(\mathbf{x}))]\right.\\ &\left.+E_{\boldsymbol{x} \sim p_{\mathbf{x}}}[\log (1-D(G(\mathbf{x})))]\right). \end{aligned}
\end{equation} 

\textbf{Image reconstruction loss:}
In order  to ``fool'' the discriminator, the generator need to generate high-quality images in the process of training. To achieve this objective, the generator $G$, including an encoder $G_{e}$ and a decoder $G_{d}$, obtains the latent feature of target class and generates high-quality images by minimizing the pixel-wise error between original input images $x$ and generated images $G(x)$. 

\begin{equation}
	\label{eq:3}
	\mathcal{L}_{irec}=\mathbb{E}_{x \sim p_{\mathbf{x}}}\|x-G(x)\|_{1}
\end{equation}

\textbf{Center distance loss: }Center distance loss could map the latent feature to center c as close as possible and make the samples from different classes more separable, which further promote the generator to capture the common factors of variation in the normal samples. Based on this, we propose a novel loss function to make the latent feature more concentrated by penalizing the smaples mapped further away from the center. Afterwards, the model could get a compact description of the  normal samples as follows.

\begin{equation}
\mathcal{L}_{c}=\min _{\mathcal{W}} \frac{1}{n} \sum_{i=1}^{n}\left\|\phi\left(\boldsymbol{x}_{i} ; \mathcal{W}\right)-\boldsymbol{c}\right\|^{2}
\end{equation}

\textbf{Latent representation loss: }Only for target class samples, the auxiliary encoder can reconstruct the latent representation z well from generated image $x'$. Besides, the latent regularizer might incur the distribution distortion in latent feature space. The feature representation $\mathbf{z'}$ can be regarded as the anchor to prevent $\mathbf{z}$ from drifting. Hence, we consider to add a constraint by minimizing the distance between latent feature of input images $G_{e}(x)$ from generator and encoded latent feature of generated image from auxiliary encoder $G_{e^{\prime}}(x^{\prime})$ as follows.

\begin{equation}
	\label{eq:4}
	\mathcal{L}_{zrec}=\mathbb{E}_{x \sim p_{\mathbf{X}}}\left\|G_{e}(x)-G_{e^{\prime}}(x^{\prime})\right\|_{2}
\end{equation}

\textbf{Dual autoencoder loss: }To obtain a more stable training process and make the samples from different class more discriminate, we regard the auxiliary autoencoder as a discriminator $D^{\prime}$. Inspired by Vu \cite{vu2019anomaly}, compared with conventional adversarial loss, we match distribution between losses, $\mathcal{L}_{direc}$ and $\mathcal{L}_{girec}$, not between samples. The training objective of this discriminator is to reconstruct the realistic inputs $x$ faithfully while fail to do so for generated input $G(x)$, as shown below:

\begin{equation}
\label{eq:5}
\mathcal{L}_{girec}=\|x-D^{\prime}(x)\|_{1},
\end{equation}

\begin{equation}
\label{eq:6}
\mathcal{L}_{direc}=\|G(x)-D^{\prime}(G(x))\|_{1},
\end{equation}

\begin{equation}
\label{eq:7}
	\mathcal{L}_{dual}=\mathcal{L}_{girec} - k\mathcal{L}_{direc},
\end{equation}where $k$ controls how much emphasis is put on the pixel-wise error of generated input $\mathcal{L}_{direc}$  during gradient descent. 

\section{Experiments}

In this section, we present the experimental setting, including datasets, evaluation measures, and more implementation details firstly. Then, ablation study are conducted to analyze the proposed method in detail. Finally, We evaluate proposed method and the state-of-the-art techniques on the same experiment setting.

\subsection{Experimental Setting}

\textbf{Datasets: }MNIST dataset includes 60,000 handwritten digits from number 0 to number 9. In order to replicate the results presented in \cite{ruff2018deep}, one of class is regarded as normal class, while the rest of the classes belong to the anomaly. In total, we get ten sets of MNIST dataset. In the same way, CIFAR10 dataset also have ten classes. We again consider one class as normal and the rest as abnormal. We then detect the anomalies by only training the model on the normal class.

 The German Traffic Sign Recognition Benchmark (GTSRB) dataset \cite{stallkamp2011german} includes adversarial boundary attack on stop signs boards.  Adversarial examples are generated from randomly drawn stop sign images of the test set using Boundary Attack Brendel \cite{brendel2017decision}, as it is done in \cite{ruff2018deep}.

\textbf{Evaluation Measures: }In the testing process, the image reconstruction loss between the realistic image and the generated image $G^{\prime}(G(x))$ is regarded as an anomaly detection score. We compute the Area Under the ROC Curve (AUC) to show the ability of the models to recognize anomalies.

\textbf{Implementation Details: }We implement our approach in PyTorch by optimizing the weighted loss $\mathcal{L}$ (defined in Eq. (\ref{loss-function}))  with the weight values $w_{i}=1$, $w_{a}=5$, $w_{z}=1$, $w_{c}=0.05$ and $w_{d}=1$, which are empirically chosen to yield optimum results. For dual autoencoder loss, the parameter k is set to 0.4. The experiments are carried out on a standard PC with a NVIDIA-1080  GPU and a multi-core 2.1 GHz CPU.

\subsection{Ablation Study}
\begin{table}[]
	\scriptsize
		\centering 
	\caption{The effect of different loss compositions is evaluated in three dataset experiments.}
	\label{tab:intra-dataset-loss}
	\begin{tabular}{|l|c|c|c|c|c|}
		\hline 
		&\multicolumn{5}{|c|}{dataset}  \\
			\hline 
		&\multicolumn{2}{|c|}{Mnist} &\multicolumn{2}{|c|}{Cifar10}&\multicolumn{1}{|c|}{GTSRB} \\
		\hline 
		Loss composition&5 &3&frog& bird&\begin{tabular}[c]{@{}l@{}}stop \\  signs \end{tabular}Í\\
		\hline 
		$\mathcal{L}_{irec}$+$\mathcal{L}_{adv}$&0.59&0.63 &0.54&0.60&0.49\\
		\hline 
		$\mathcal{L}_{irec}$+$\mathcal{L}_{adv}$+$\mathcal{L}_{zrec}$&0.63&0.68&0.58&0.61&0.50\\
		\hline 
		${L}_{irec}$+$\mathcal{L}_{adv}$+$\mathcal{L}_{zrec}$+$\mathcal{L}_{c}$&0.65&0.70&0.60&0.63&0.55\\
		\hline 
		${L}_{irec}$+$\mathcal{L}_{adv}$+$\mathcal{L}_{zrec}$+$\mathcal{L}_{c}$+$\mathcal{L}_{d}$&0.99&0.99&0.97&0.98&0.96\\
		\hline 
	\end{tabular}	
\end{table}

\begin{figure} 
	\small
	\centering 
	\includegraphics[width=\linewidth]{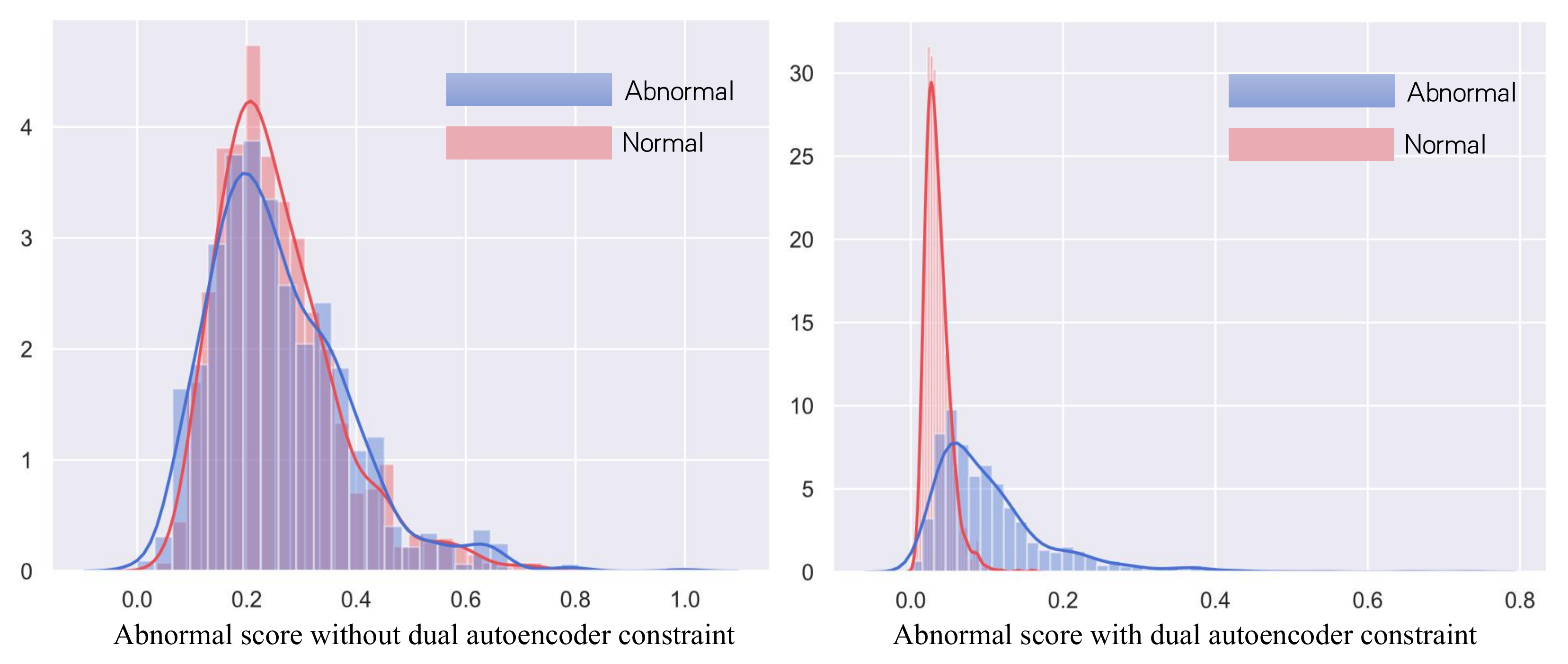} 
	\caption{Histogram of the scores for both normal and abnormal test samples.}
	\label{his} 
\end{figure}

\begin{table}[]
	\scriptsize
	\centering 
	\caption{Average AUCs in \% with StdDevs (over 10 seeds) per method on GTSRB stop signs with adversarial attacks.}
	\label{stop}
	\begin{tabular}{ll}
		\hline
		\textbf{Methods} & \textbf{Accuracy} \\
		\hline
		
		OC-SVM/SVDD            & 67.5$\pm$1.2        \\
		KDE           & 60.5$\pm$1.7           \\
		IF           & 73.8$\pm$0.9           \\
		DCAE              & 79.1$\pm$3.0        \\
		ANOGAN          &     $-$      \\
		SOFT-BOUND. DEEP SVDD            & 77.8$\pm$4.9            \\
		ONE-CLASS DEEP SVDD        & 80.3$\pm$2.8          \\
		RCAE        & 87.4$\pm$2.7          \\
		\textbf{Ours}    &\textbf{96.9$\pm$1.0}    \\
		\hline
	\end{tabular}
\end{table}

\begin{figure} 
	\small
	\centering 
	\includegraphics[width=\linewidth]{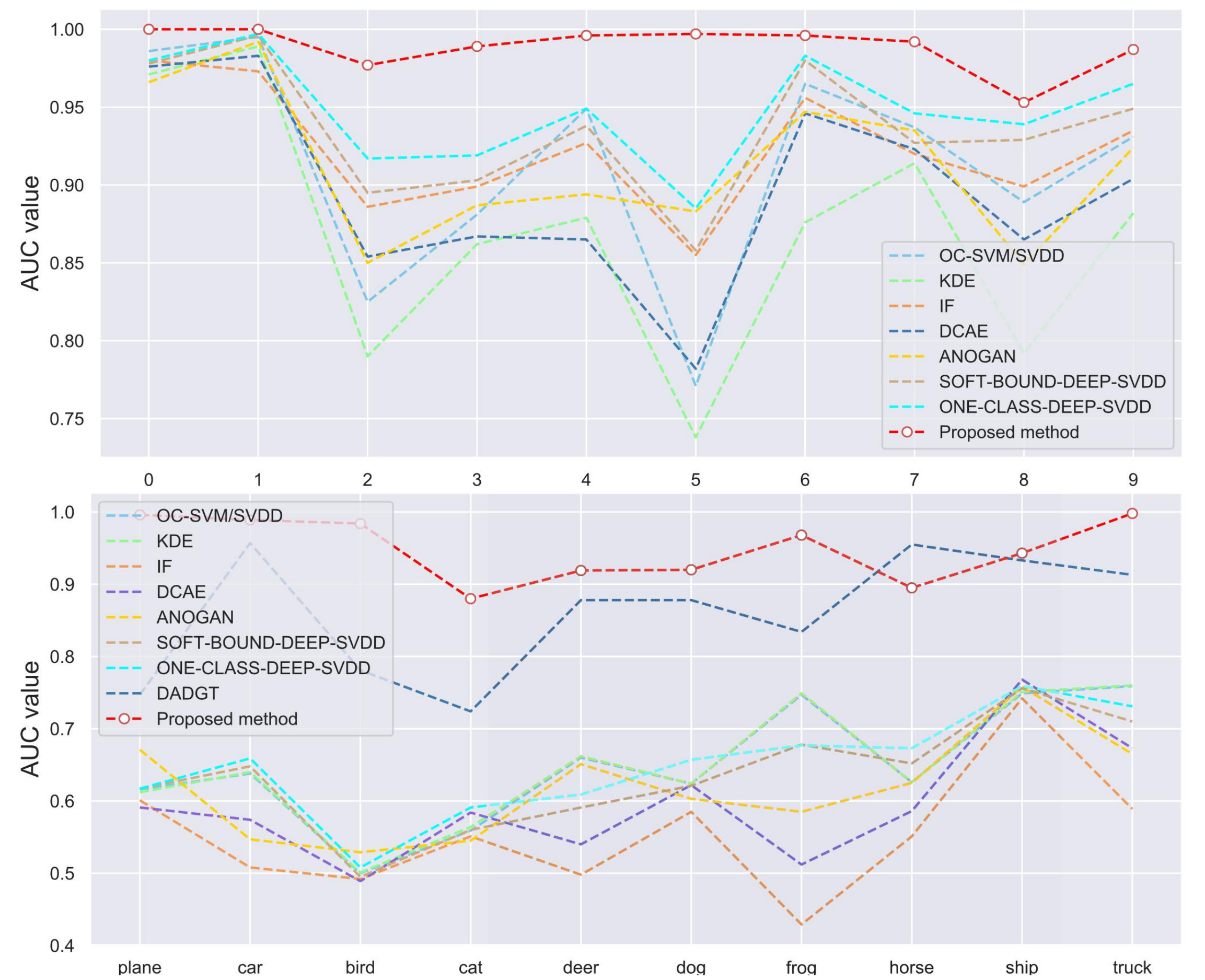} 
	\caption{Digit and class designated as normal class in MNIST (Top) and CIFAR10 dataset (Bottom) }
	\label{auc_mnist_cifar} 
\end{figure}

Since the proposed architecture is built upon the adversarial training framework, recall the subsection 2.2, it is necessary for us to give the ablation study to present the different combinations of loss functions on three different datasets. As illustrated in Tab. \ref{tab:intra-dataset-loss}, generally speaking, all of proposed loss functions we designed in our framework can improve the performance. To further understand the importance of dual autoencoder constraint, we visualize  the distribution of abnormal scores from 200 random samples to make an intuitive explanation, as illustrated in Fig. \ref{his}, where the proposed method with and without constraint are compared on horse experiment setting. To sum up, thanks to the dual autoencoder constraint, we can easily observe that not only the distribution of normal or abnormal samples are more concentrated, but also the distribution of anomalous samples can be separated from that of normal samples.

\subsection{Comparison with State-of-the-art Methods}

In the stop sign dataset, proposed method exactly detect the adversarial boundary attack and achieve the best performance in Tab. \ref{stop}. Besides, we also present that proposed method has the clear superiority over cutting-edge abnormal detectors, as it is shown in Fig. \ref{auc_mnist_cifar}. For MNIST dataset, we select one class as the normal each time, while leaving the rest to be the anomaly, leading to ten sets for abnormal detection. Our method achieves significantly improvements compared with other compatitors, in term of AUC value, which could be illustrated by using red curves in Fig. \ref{auc_mnist_cifar}. Except for MNIST, we also present the comparison in CIFAR10 dataset.

\section{Conclusion}
To exploit more separable latent representation features between the normal samples and the anomalies in a stable training process, this paper proposed a novel dual adversarial network, in which the underlying structure of training data is not only captured in latent feature space, but also can be further restricted in the space of latent representation in a discriminant manner. In addition, to tackle with an imbalance of capability between generator and discriminator which results in unstable training and hinders the ability of GAN, the auxiliary autoencoder is employed as a discriminator, leading to a better balance in the process of training. Extensive experiments have been conducted on some datasets, showing high performance of proposed method and the benefit of using the latent regularizer and the auxiliary autoencoder.

\clearpage
\bibliographystyle{IEEEbib}
\bibliography{refs}

\end{document}